\documentclass[10pt,twocolumn,letterpaper]{article}

\usepackage{iccv}              
\usepackage[accsupp]{axessibility}  

\definecolor{iccvblue}{rgb}{0.21,0.49,0.74}
\usepackage[pagebackref,breaklinks,colorlinks,allcolors=iccvblue]{hyperref}
\usepackage{colortbl}
\usepackage{wrapfig}

\usepackage{graphicx}
\usepackage{animate}
\usepackage{makecell}

\def\paperID{8933} 
\def\confName{ICCV}
\def\confYear{2025}

\title{Omegance: A Single Parameter for Various Granularities in \\ Diffusion-Based Synthesis}

\author{Xinyu Hou \qquad Zongsheng Yue \qquad Xiaoming Li \qquad Chen Change Loy\\
S-Lab, Nanyang Technological University\\
{\tt\small xinyu.hou@ntu.edu.sg\qquad zsyzam@gmail.com\qquad csxmli@gmail.com\qquad
 ccloy@ntu.edu.sg}}

\begin{document}

\twocolumn[{%
\renewcommand\twocolumn[1][]{#1}%
\maketitle
\vspace{-10mm}
\thispagestyle{empty}
\begin{center}
    \centering
    \includegraphics[width=0.93\linewidth]{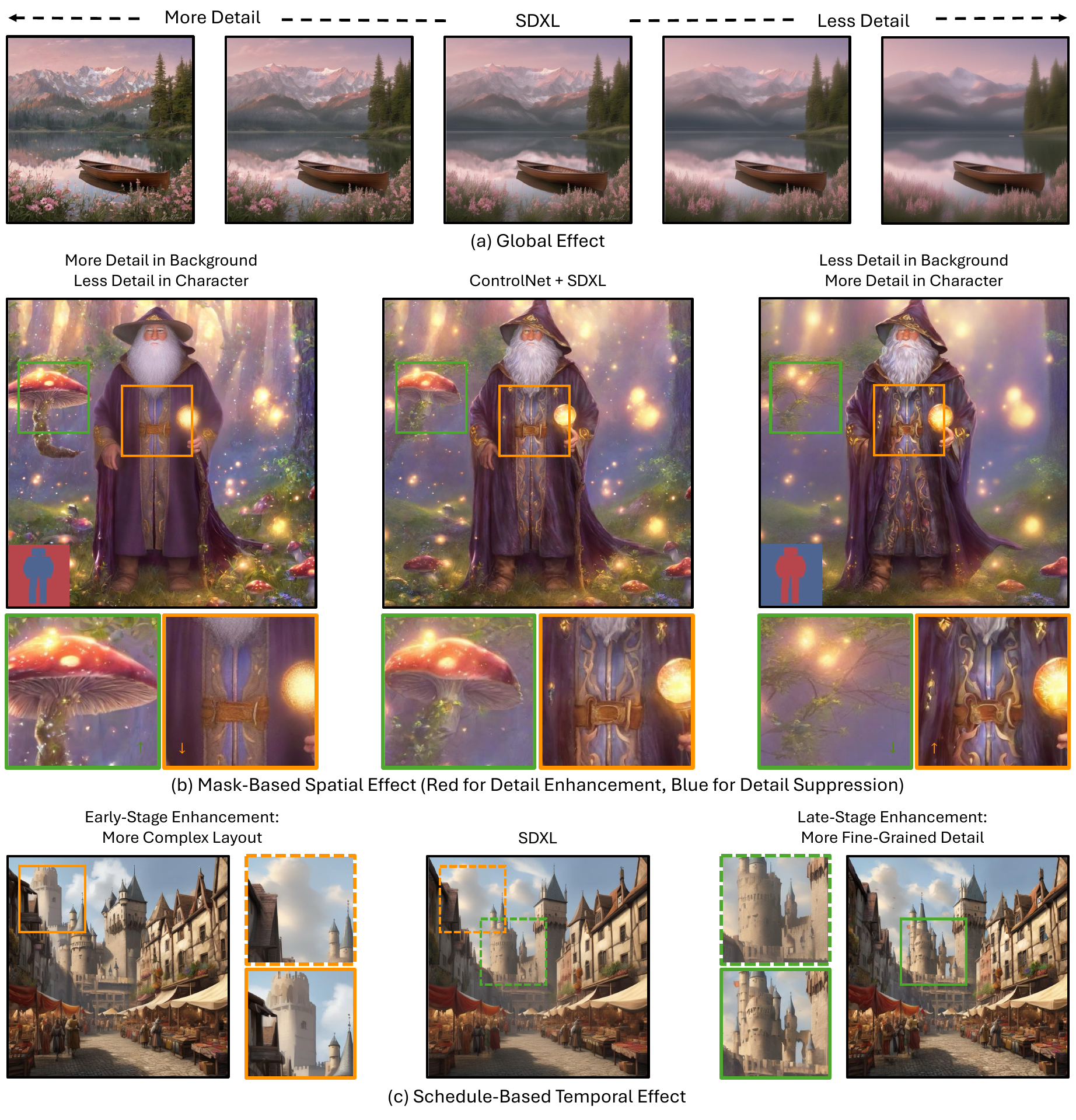}
    \vspace{-2mm}
    \captionof{figure}{
    \textbf{Omegance} enables flexible granularity control over generation results. The control can be implemented globally, spatially with an omega mask, or temporally with an omega schedule.
    \textit{(Zoom-in for best view)}
    } \vspace{1mm}
    \label{fig:teaser}
\end{center}%
}]

\begin{abstract}
    In this work, we show that we only need \textbf{a single parameter $\omega$} to effectively control granularity in diffusion-based synthesis. This parameter is incorporated during the denoising steps of the diffusion model’s reverse process. This simple approach does not require model retraining or architectural modifications and \textcolor{black}{incurs negligible computational overhead}, yet enables precise control over the level of details in the generated outputs. Moreover, spatial masks or denoising schedules with varying $\omega$ values can be applied to achieve region-specific or timestep-specific granularity control. 
    \textcolor{black}{External control signals or reference images can guide the creation of precise $\omega$ masks, allowing targeted granularity adjustments.}
    Despite its simplicity, the method demonstrates impressive performance across various image and video synthesis tasks and is adaptable to advanced diffusion models. The code is available at \url{https://github.com/itsmag11/Omegance}.
    
\end{abstract}    
\section{Introduction}
\label{sec:intro}

Diffusion models have emerged as powerful tools in image and art generation by progressively transforming random noise into coherent visual content through a learned iterative process~\cite{ho2020ddpm, podell2024sdxl, saharia2022photorealistic, midjourney, baldridge2024imagen3, ho2022imagenvideo, li2023w-plus-adapter}. 
\textcolor{black}{They have become the dominant paradigm for high-quality image synthesis, offering strong diversity and controllability.}

Artists and designers often need to strategically decide where and how to apply details in their work. The level of detail in a piece of artwork or photograph can shape its visual harmony, order, and clarity, influencing how the viewer experiences and interprets it while guiding their focus~\cite{arnheim_art_2020, followingthemasters}.
The vanilla diffusion model does not inherently offer direct, fine-tuned control over the level of granularity in specific areas of an image. While the model can generate varying levels of detail across different images, its uniform generative process does not allow for easy manipulation of how much detail is rendered in different parts of the same image. The level of detail in an image can be challenging—or even impossible—to convey through text alone. For instance, reducing detail in the background while retaining high detail in the main subject (see the right case of Fig.~\ref{fig:teaser}(b) for illustration) is not straightforward.

In this paper, we explore a novel yet simple approach for controlling the level of detail in diffusion model outputs by scaling the predicted noise during each denoising step. The method does not require any network architecture or noise scheduling modifications. Instead, we demonstrate that it can influence the granularity of the visual output by dynamically adjusting the variance of the removed noise at each step. 
\textcolor{black}{While variance scaling is a fundamental operation in diffusion models, to the best of our knowledge, it has not been systematically explored as a means for fine-grained granularity control.}
This simple yet flexible technique allows tailored adjustments in concept density and object texture, offering users a more nuanced control over the synthesized content.

\textcolor{black}{
The presented approach is named \textbf{Omegance}, combining ``omega'' and ``nuance''. It is appealing as it enables noise scaling with a single parameter, $\omega$.}
Decreasing $\omega$ results in less noise being removed, leading the network to infer more complex scenes and richer textures. Conversely, increasing $\omega$ removes more noise, leading to smoother and simpler outputs.
While applying our omega control globally over space and consistently over time can yield uniformly richer or smoother results, as shown in Fig.~\ref{fig:teaser}(a), more precise controls can be implemented both spatially and temporally. (1) Since the granularity requirement may vary within a single image, e.g., finer-grained details for areas requiring rich textures and complex visual elements, and coarser-grained details for areas demanding smooth transitions and high-level quality, we can use \textbf{omega masks} to customize the desired effects across different spatial regions. Examples of different spatial effects are shown in Fig.~\ref{fig:teaser}(b). The mask can be created either from user-provided strokes or generated using specific guiding conditions. (2) To better align with the diffusion denoising dynamics~\cite{ho2020ddpm, song2021scorebased}, where the object shapes and image layouts typically emerge in the early stages and fine details in the later stages, we can implement \textbf{omega schedules}, adjusting the omega value over time for varying effects on layout and detailed textures. Examples are shown in Fig.~\ref{fig:teaser}(c).

Omegance is not limited to any specific network architecture or denoising scheduler as long as the progressive diffusion denoising process is followed. Extensive experiments demonstrate Omegance's ability to adapt to various diffusion-based synthesis tasks. Models evaluated include Stable Diffusion~\cite{podell2024sdxl, esser2024scaling} and FLUX~\cite{flux_github} for text-to-image generation, SDEdit~\cite{meng2022sdedit} and ControlNet~\cite{zhang2023controlnet} for image-to-image generation, SDXL-Inpainting~\cite{podell2024sdxl} for image inpainting, ReNoise~\cite{garibi2024renoise} for real-image editing, and Latte~\cite{ma2024latte} and AnimateDiff~\cite{guo2023animatediff} for text-to-video generation. Some examples are shown in Fig.~\ref{fig:teaser}. In all of the above applications, effective and smooth, nuanced control over the generated results is observed, demonstrating the effectiveness of our single-parameter granularity adjustment. 

\section{Related Work}
\label{sec:related_work}

\noindent\textbf{Diffusion-based Editing.}
Most previous diffusion-based editing methods focus on exploiting the visual-language association ability of CLIP~\cite{Radford2021clip} to edit visual content according to language guidance. Prompt-to-Prompt~\cite{hertz2023prompttoprompt} and InstructPix2Pix~\cite{brooks2022instructpix2pix} edit concepts in the output by modifying the cross-attention maps, which play a crucial role in aligning textual prompts with visual features during the generation process. SEGA~\cite{brack2023sega} generates results following the semantic guidance of a target prompt during denoising. Wu~\etal~\cite{wu2023disentanglementdiffusion} find by mixing the text embedding of prompts with and without the target attribute, the output can preserve the original content while aligning with the desired attribute. Besides, SDEdit~\cite{meng2022sdedit} adds noise to the modified image and utilizes diffusion prior to rationalize the edited parts as natural images. 
\textcolor{black}{These methods struggle when the desired edits cannot be clearly described using language or inferred from the original image}
, and fail to provide a flexible way to edit the granularity of the output.

\noindent\textbf{Generation Quality Enhancement.}
Efforts have also been made to enhance the quality of content generated by diffusion models. Several works have explored modifications of Classifier-Free Guidance (CFG)~\cite{ho2021classifierfree} to improve generation quality~\cite{hong2023sag, ahn2024pag, sadat2024notrainingnoproblem}. SAG~\cite{hong2023sag} and PAG~\cite{ahn2024pag} substitute the null-text prediction in CFG with self-attention or perturbed self-attention maps, enabling high-quality, training- and condition-free generation. Sadat~\etal~\cite{sadat2024notrainingnoproblem} present a guidance strategy similar to CFG, applied between clean and perturbed text embeddings, to boost generation quality. 
While these methods effectively enhance quality globally, they lack the capability for spatially fine-grained control over details in the generated output~\cite{ho2020ddpm,song2021ddim,karras2022elucidating,nichol2021improved}. 

Another line of research leverages reinforcement learning from human feedback (RLHF)~\cite{ouyang2022training, rafailov2023direct, christiano2017deep, schulman2017proximal} to fine-tune diffusion models for higher-quality results aligned with human preferences~\cite{xu2023imagereward, 2023DPOK, yang2023using}. Xu~\etal~\cite{xu2023imagereward} present a general-purpose text-to-image human preference reward model and use it to fine-tune the diffusion model regarding human preference score. A similar approach is adopted in concurrent work DPOK~\cite{2023DPOK}. Furthermore, Yang~\etal~\cite{yang2023using} employ direct preference optimization (DPO), fine-tuning the diffusion model to align with human feedback without a separate reward model. While these methods produce outputs that reflect human preferences, they involve costly model fine-tuning and lack flexible control over output granularity. 
FreeU~\cite{si2023freeu} was recently introduced to enhance the quality of diffusion model outputs, specifically targeting the U-Net architecture in the denoising process. This method involves jointly adjusting two scaling factors during inference: one for amplifying the backbone features and one for modulating the influence of the skip connections to better preserve details without over-smoothing or degrading high-frequency elements. 
While achieving noticeable quality improvement, FreeU is closely tied to the U-Net architecture and requires careful adjustment of its dual scaling parameters.
In contrast, Omegance provides a simpler, more flexible, and architecture-agnostic approach to control the level of detail in diffusion models.

\noindent\textbf{Noise Scheduling.}
\textcolor{black}{Noise scheduling is also crucial for diffusion model performance, impacting generation quality and stability. The linear and cosine schedulers~\cite{nichol2021improved} are widely used, with the linear scheduler applying uniform noise variance and the cosine scheduler preserving high-frequency details longer. Lin et al.~\cite{lin2024common} identify flaws for not enforcing zero terminal SNR in previous schedulers and propose a rescaled scheduler, correcting noise variance scaling for improved stability and fidelity. These findings underscore the need for careful noise schedule design to enhance sample quality.
Unlike traditional noise schedulers that apply fixed, global denoising adjustments, which can be unpredictable and require model retraining, Omegance offers a lightweight, interpretable, and architecture-agnostic mechanism for both global and localized detail control, seamlessly integrating with existing schedulers.}
\section{Methodology}
\label{sec:method}

\begin{figure*}[h]
    \centering
    \includegraphics[width=0.95\linewidth]{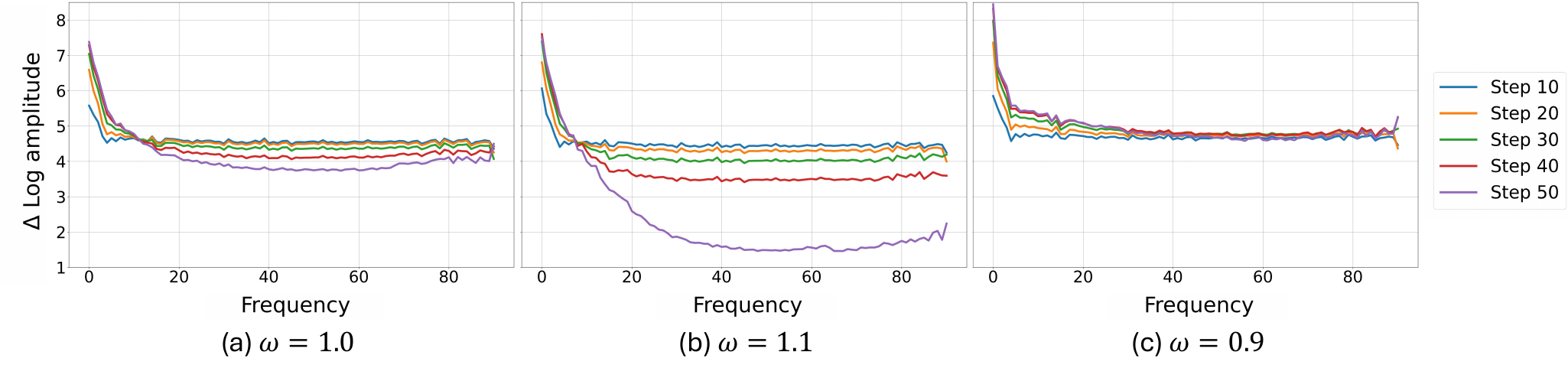}
    \vspace{-4mm}
    \caption{Effects of Omegance on the frequency spectrum of the intermediate latent $z'_t$. The inference step in the legend is inversely correlated with the timestep $t$. In the original denoising process (a), high-frequency components gradually diminish while low-frequency components become more prominent as the denoising progresses to late stages. With Omegance, increasing $\omega$ leads to more aggressive removal of the high-frequency components, as shown in (b), and vice versa, as depicted in (c).}
    \label{fig:freq}
    \vspace{-4mm}
\end{figure*}

\subsection{Diffusion Model Preliminaries}
\label{sec:pre}

Diffusion models are powerful generative models that synthesize realistic images by iteratively predicting the noise added to a sample. They consist of two processes: In the forward process, Gaussian noise $\epsilon$ is progressively added to an initial latent $z_0$ that is directly decoded from $x_0$. Following Song~\etal~\cite{song2021ddim}, we formulate the process as:
\begin{equation}
    z_t = \sqrt{\alpha_t} z_0 + \sqrt{1 - \alpha_t} \epsilon, \quad \epsilon \sim \mathcal{N}(0, 1)
    \label{equ:forward}
\end{equation}
\noindent where $z_t$ is the noisy latent at timestep $t$. The noise schedule $\alpha_t$ is defined as the cumulative product of $(1 - \beta_t)$: $\alpha_t = \prod_{i=1}^{t} (1 - \beta_i)$ where $\beta_t$ is a predefined variance schedule that controls the amount of noise added at each timestep~\cite{ho2020ddpm}. \footnote{Note that some papers~\cite{ho2020ddpm, nichol2021improved, lin2024common} denote $(1 - \beta_t)$ as $\alpha_t$ and $\prod_{i=1}^{t} \alpha_t$ as $\bar{\alpha_t}$.} In the reverse process, pure Gaussian noise is transformed into coherent visual content through a learned denoising process. A general representation of the denoising process is:
\begin{equation}
    z_{t-1} = \delta_t \cdot z_t +  \underbrace{\zeta_t \cdot \epsilon_\theta(z_t, t)}_{\text{``direction pointing to $z_0$''}}
    \label{equ:denoise}
\end{equation}
\noindent where $\delta_t$ and $\zeta_t$ are scaling factors of the current noisy signal and the noise prediction, respectively, that vary according to specific scheduler \textit{(see supp.~Sec.A for details)}, and $\epsilon_\theta (z_t, t)$ is the noise prediction at time $t$ by the denoising network with parameters $\theta$.  This formula characterizes the iterative denoising process, where each step aims to progressively move from a noisier latent $z_t$ towards clean $z_0$ to get a less noisy latent $z_{t-1}$.

\if
During each denoising step, the previous noisy sample is obtained by:
\begin{equation}
    \begin{split}
        & x_{t-1} = \\
        & \sqrt{\alpha_{t-1}} \left( \frac{x_t - \sqrt{1 - \alpha_t} \epsilon_\theta (x_t, t)}{\sqrt{\alpha_t}} \right) + \sqrt{1 - \alpha_{t-1}} \epsilon_\theta (x_t, t)
    \end{split}
\end{equation}
\noindent where $\epsilon_\theta (x_t, t)$ is the noise prediction at time $t$ by the denoising network with parameters $\theta$. Here, $\alpha_t$ helps the model scale and balance between the denoised prediction and the residual noise at each timestep, allowing it to iteratively reconstruct the image by gradually reducing noise.
\fi

\noindent
\textbf{Signal-to-Noise Ratio.} In diffusion models, the Signal-to-Noise Ratio (SNR) plays a critical role in defining the balance between the original image content $z_0$ and added Gaussian noise $\epsilon$ at each timestep. Given Equ.~\eqref{equ:forward}, $\mathrm{SNR}$ is defined as:
\begin{equation}
    \mathrm{SNR}(t) = \frac{\alpha_t}{1 - \alpha_t}.
\end{equation}
\noindent \textcolor{black}{When $t \rightarrow 0$, $\mathrm{SNR} \rightarrow \infty$ indicating pure image signal $z_0$.} As $t$ increases, SNR decreases to $0$, demonstrating pure noise $z_T$. 
During denoising, 
the model progressively aligns the SNR of each timestep with the SNR defined by the forward process.
Since $\alpha_t$ is predefined by the noise schedule, the SNR in vanilla diffusion models remains fixed throughout the denoising process, limiting flexibility in controlling the amount of noise at each timestep.

\noindent
\textbf{Denoising Dynamics.} In diffusion models, denoising dynamics follow a progressive refinement~\cite{ho2020ddpm, song2021scorebased}: broad structures, such as image layout and object shapes, emerge in early stages, while fine-grained details appear in later steps. This behavior reflects the nature of the forward diffusion process, where high-frequency details are corrupted by noise first and low-frequency broad structures last, resulting in an inverse reconstruction during denoising. Such dynamics not only stabilize generation but also enable flexible, hierarchical control over image features at different timesteps.

\begin{figure*}[t]
    \centering
    \includegraphics[width=0.95\linewidth]{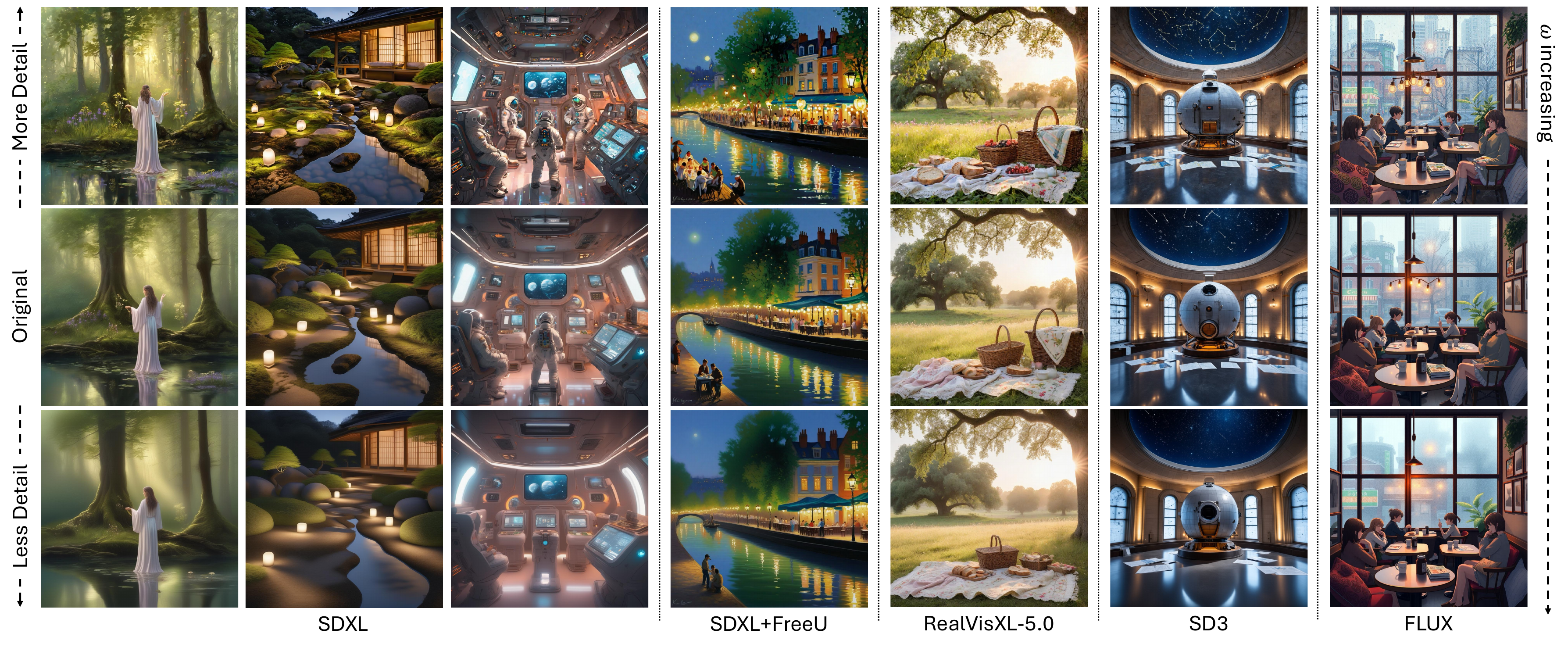}
    \vspace{-4mm}
    \caption{Global effects of Omegance. The models indicated below are the base models. The middle row shows the original base model results. The top and bottom rows are Omegance results with detail enhancement and suppression, respectively. Omegance can effectively add or remove details without harming visual quality or modify the entire image completely, making it a flexible tool for practical use. 
    }
    \label{fig:global}
    \vspace{-4mm}
\end{figure*}

\subsection{Omegance}
\label{sec:omegance}

We introduce Omegance, which uses a parameter $\omega$ to scale the noise prediction at each denoising step in the reverse diffusion step. A general form of a single denoising step with Omegance is formulated as follows:
\begin{equation}
    z_{t-1}' = \delta_t \cdot z_t + \underbrace{\zeta_t \cdot \epsilon_\theta(z_t, t) \cdot \omega}_{\text{``modified direction pointing to $z_0$''}}
    \label{equ:omegance}
\end{equation}
\noindent \textcolor{black}{Since the diffusion model is trained to predict Gaussian noise, $\epsilon_\theta(z_t, t)$ can be viewed as an estimation of the standard Gaussian noise $\epsilon \sim \mathcal{N}(0, 1)$. While $\epsilon_\theta$ itself is a deterministic output and not necessarily Gaussian-distributed, scaling it by a factor $\omega$ still serves to control the effective denoising direction during sampling. This controls how much detail is recovered, without altering the underlying direction toward the clean signal.}
\textcolor{black}{In practice, $\omega$ is rescaled by $\omega = \mathcal{R}(\varpi)$ to allow input $\varpi \in (-\infty, \infty)$ for finer-grained control and re-centered at $0$ \textit{(see formulation and sensitivity test in supp.~Sec.J.3)}.}

Though it is a simple introduction of a coefficient to the denoising term, its influence on SNR and detail generation is worth investigating. Taking DDIM scheduler~\cite{song2021ddim} as an example, the modified SNR during denoising is formulated as follows \textit{(see supp.~Sec.C for step-by-step derivations)}:
\begin{equation}
\begin{split}
    &\mathrm{SNR}(t-1)' = \\ 
    &\frac{\alpha_{t-1}}{\left[ \frac{\sqrt{\alpha_{t-1}} \sqrt{1 - \alpha_t}}{\sqrt{\alpha_t}} + \omega \left(\frac{\sqrt{\alpha_{t}}\sqrt{1 - \alpha_{t-1}} - \sqrt{\alpha_{t-1}} \sqrt{1 - \alpha_t}}{\sqrt{\alpha_t}} \right)\right]^2}
\end{split}
\label{equ:mod_snr}
\end{equation}
\noindent where $\sqrt{\alpha_{t}}\sqrt{1 - \alpha_{t-1}} - \sqrt{\alpha_{t-1}} \sqrt{1 - \alpha_t}$ is always negative due to the monotonically-decreasing nature of $\alpha_t$.

\begin{itemize}
\item When $\omega = 1$, $\mathrm{SNR}(t-1)' = \mathrm{SNR}(t-1)$.
Omegance retains the standard denoising schedule as in Equ.~\eqref{equ:denoise}, leaving the amount of noise removed from $z_t$ unchanged. The SNR schedule aligns with the forward process. This setting produces a balanced output with standard levels of detail and texture across the entire image, aligning with the expected granularity of the original noise schedule. 

\item When $\omega < 1$, $\mathrm{SNR}(t-1)' < \mathrm{SNR}(t-1)$.
The noise prediction is scaled down, leading to a less aggressive denoising towards $z_0$.
Therefore, the latent state $z'_{t-1}$ retains additional high-frequency information, as illustrated in Fig.~\ref{fig:freq}(c).
With the noise component dominating, the model ``justifies'' this residual noise by generating more intricate structures and richer textures, enhancing visual complexity in the output.

\item When $\omega > 1$, $\mathrm{SNR}(t-1)' > \mathrm{SNR}(t-1)$.
The denoising schedule becomes more aggressive. This amplified noise reduction diminishes high-frequency information in the latent $z'_{t-1}$. With the signal now dominating, the model interprets the reduced residual noise as a cue to simplify textures and details, yielding smoother and less intricate visual outputs.
\end{itemize}

\noindent
Both rich and smooth effects can be desirable depending on the user's intent. For instance, setting $\omega < 1$ enhances detail, making it well-suited for generating a busier crowd in a marketplace, intricate patterns in clothing design, or fine textures in elements like sand or waves. On the other hand, $\omega > 1$ produces smoother, simpler visuals, ideal for scenes with clear skies, calm waters, or minimalist designs, where a streamlined aesthetic is preferred. This flexibility allows users to adapt granularity dynamically to match specific visual and stylistic goals.


\noindent
\textbf{Omegance in Various Schedulers.} Omegance can be applied in various noise schedulers. Below, we outline the modified denoising step formula for several popular schedulers. For DDIM~\cite{song2021ddim} and Euler discrete~\cite{Karras2022edm} schedulers where the standard noise added in the current step $\epsilon_\theta(z_t, t)$ is available (in Euler scheduler, it approximate the ``derivative'' of $z$), we can directly apply $\omega$ on it to achieve mean-preserving variance modification.

\noindent (1) DDIM scheduler~\cite{song2021ddim}:
\begin{equation}
    \begin{split}
    z_{t-1}' &= \sqrt{\alpha_{t-1}} \left( \frac{z_t - \sqrt{1 - \alpha_t} \cdot \epsilon_\theta(z_t, t) \cdot \boldsymbol{\omega}}{\sqrt{\alpha_t}} \right) \\
    &+ \sqrt{1 - \alpha_{t-1}} \cdot \epsilon_\theta(z_t, t) \cdot \boldsymbol{\omega}
    \end{split}
\end{equation}

\noindent (2) Euler discrete scheduler~\cite{Karras2022edm}:
\begin{equation}
    z_{t-1}'
    = z_t + (\sigma_{t+1} - \hat{\sigma})\cdot \epsilon_\theta(z_t, t) \cdot \boldsymbol{\omega}
\end{equation}
\noindent where $\sigma$ is the noise level from Karras~\etal~\cite{Karras2022edm}, and $\hat{\sigma} = \sigma_t \cdot (\gamma + 1)$ when $\gamma$ is the ``churn'' factor for perturbing $\sigma_t$.

However, in the flow-matching-based scheduler~\cite{esser2024scaling}, the forward process learns a continuous transformation without the need for a stepwise noise addition schedule: $z_t = (1 - t) z_0 + t\epsilon$, where $\epsilon \sim \mathcal{N}(0, 1)$, which slightly differs from Equ.~\eqref{equ:forward}. During the reverse process, the model predicts $v_\theta(z_t, t) = \frac{dz_t}{dt} = \epsilon - z_0$, and moves one step forward with $z_{t-1} = z_t + dt \cdot v_\theta(z_t, t)$. Here, the term $dt \cdot v_\theta(z_t, t)$ represents the denoise amount as in the general formula Equ.~\eqref{equ:denoise}, but is not necessarily a standard noise. To prevent mean-shifting \textcolor{black}{which causes unwanted color change \textit{(details in supp.~Sec.I.2)}}, we apply a mean-preserving operation with Omegance in the flow matching scheduler.

\noindent (3) Flow matching scheduler~\cite{esser2024scaling}:
\begin{equation}
\begin{split}
    m &= \mathbb{E}[dt \cdot v_\theta(z_t, t)] \\
    z_{t-dt}' &= z_t + [(dt \cdot v_\theta(z_t, t) - m) \cdot \boldsymbol{\omega} + m]
\end{split}
\end{equation}

\begin{figure}[t]
    \centering
    \includegraphics[width=0.85\linewidth]{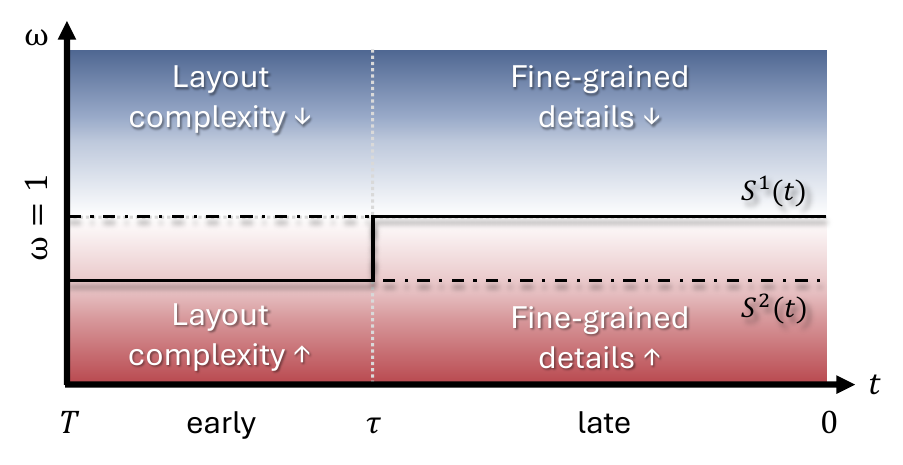}
    \vspace{-4mm}
    \caption{Illustration of $\omega$ effect during denosing process. During the early stage ($t\in[T,\tau]$), a higher $\omega$ reduces layout complexity (blue region), while a lower $\omega$ enhances it (red region). In the late stage ($t\in[\tau,0]$), a higher $\omega$ suppresses fine-grained details, whereas a lower $\omega$ enhances them. The $\mathcal{S}^1(t)$ and $\mathcal{S}^2(t)$ schedules correspond to Early-Stage Enhancement (left) and Late-Stage Enhancement (right) cases in Fig.~\ref{fig:teaser}(c). More examples in Fig.~\ref{fig:more_schedules}.
    }
    \label{fig:omega-schedule}
    \vspace{-4mm}
\end{figure}

\subsubsection{Omega Mask}
The omega mask $\omega_{i, j} = \mathcal{M}(i, j)$ introduces a spatially varying control over the granularity within a single image by allowing different regions to have distinct $\omega$ values during the denoising process.
$\mathcal{M}$ is a mask $\in \mathbb{R}^{H' \times W'}$ where $H' = H / f, W' = W / f$ are the original image dimension $H, W$ scaled by the VAE downsampling factor $f$. The mask can be obtained from user-provided strokes, segmentation masks, or automatically generated from control signals like pose skeleton, depth map, \etc. in both discrete and continuous manners as illustrated in Fig.~\ref{fig:controlnet}. 
This spatial control leverages the locality of the denoising process, ensuring that adjustments to $\omega$ in one region do not affect the $\mathrm{SNR}$ or visual properties of neighboring areas. Such flexibility is valuable for applications requiring region-specific detail control within a single image, enabling fine-grained textures in focal regions while maintaining smoothness elsewhere. 


\subsubsection{Omega Schedule}
\begin{figure}[t]
    \centering
    \includegraphics[width=\linewidth]{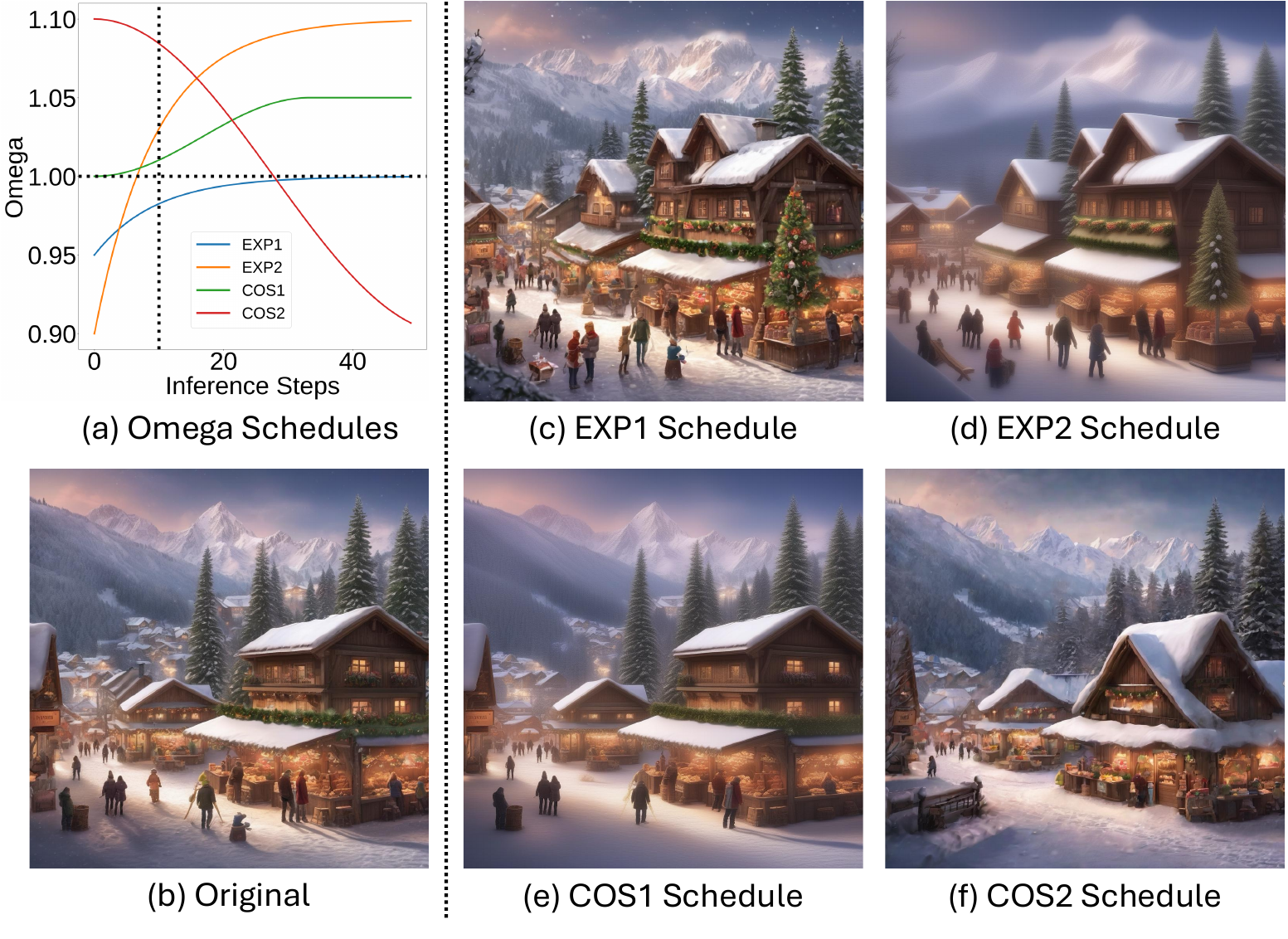}
    \vspace{-6mm}
    \caption{Temporal effects of schedule-based Omegance. Four quadrants in (a) are the same as those in Fig.~\ref{fig:omega-schedule}. Four examples are illustrated: \textbf{EXP1}: More complex layout, slightly more fine detail. \textbf{EXP2}: More complex layout, less fine detail. \textbf{COS1}: Slightly less complex layout, less fine detail. \textbf{COS2}: Less complex layout, more fine detail. \textit{(See implementation details in supp. Sec.D)}
    }
    \label{fig:more_schedules}
    \vspace{-4mm}
\end{figure}

The omega schedule $\omega_t = \mathcal{S}(t)$ provides a mechanism for controlling granularity across different stages of the denoising process by dynamically adjusting $\omega$ values over time. By introducing $\omega$ at specific stages in the reverse diffusion process, the omega schedule allows targeted influence on both the broad layout and fine-grained details within the generated image. This temporal control is aligned with the denoising dynamics: early denoising stages primarily reconstruct the general structure and layout, while later stages refine finer details. 
The effects of applying omega in different denoising stages are illustrated in Fig.~\ref{fig:omega-schedule}. 
Note that the early stage for layout formation occupies only a small portion of the overall schedule, typically within the first 10 steps in a 50-step denoising process ($\tau \approx 10$ when $T=50$), due to the fact that layout information is only corrupted in the last few steps of the forward process. 
\textcolor{black}{More schedules and their effects are visualized in Fig.~\ref{fig:more_schedules}}.
The omega schedule enables stage-specific control over image synthesis, allowing for nuanced manipulation of both composition and detail. This flexibility supports a range of creative and practical applications where different stages of the denoising process demand distinct levels of control.





\begin{figure}[t]
    \centering
    \vspace{-3mm}
    \includegraphics[width=0.9\linewidth]{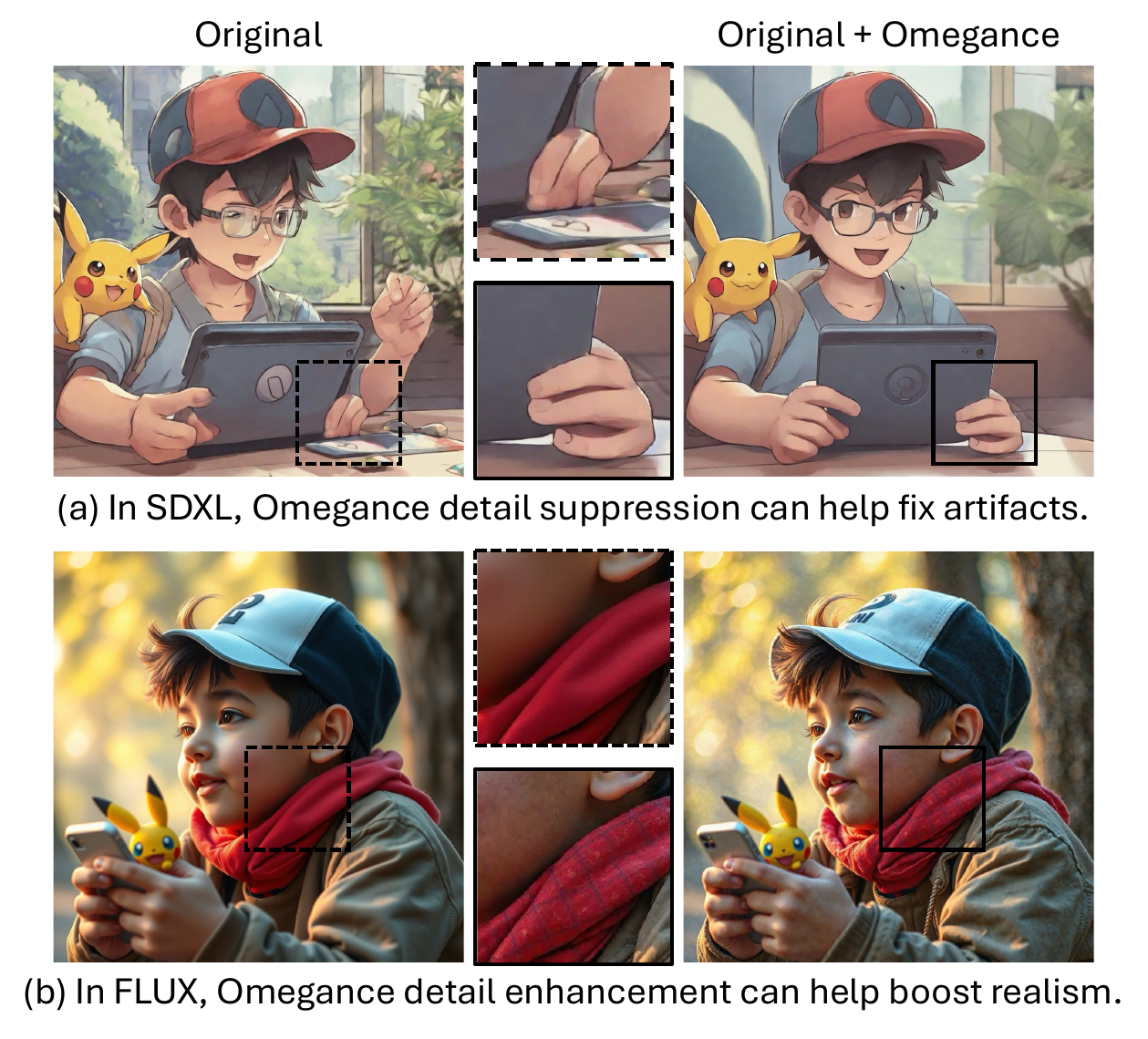}
    \vspace{-3mm}
    \caption{The effects of global Omegance in fixing visual artifacts and improving realism. 
    }
    \label{fig:quality}
    \vspace{-4mm}
\end{figure}

\section{Experiments}

We examine the effectiveness of Omegance across various generative models 
and applications, including Stable Diffusion XL (SDXL)~\cite{podell2024sdxl}, RealVisXL-V5.0~\cite{realvisxl5_huggingface}, Stable Diffusion 3 (SD3)~\cite{esser2024scaling}, FLUX~\cite{flux_github}, FreeU~\cite{si2023freeu}, SDEdit~\cite{meng2022sdedit}, ControlNet~\cite{zhang2023controlnet}, ReNoise~\cite{garibi2024renoise}, SDXL-Inpainting~\cite{podell2024sdxl}, Latte~\cite{ma2024latte}, and AnimateDiff~\cite{guo2023animatediff}. Implementations of these methods are based on Huggingface's Diffusers repository\footnote{https://github.com/huggingface/diffusers}. 
More implementation details are in supp.~Sec.J.



\subsection{Text-to-Image Generation}

\begin{table*}[t]
    \centering
    
    \vspace{-2mm}
\end{table*}

\begin{table*}[ht]
    \parbox{.65\linewidth}{
        \centering
        \footnotesize
        \caption{\textcolor{black}{Quantitative comparisons of image quality, text-image alignment, and aesthetics with previous works. Different Omegance settings are highlighted by a blue background. \textbf{Bold} and \underline{underline} indicate best and second best results, respectively.}}
        \begin{tabular}{cccccc}
        \toprule
     & FID$\downarrow$ & IS$\uparrow$ & CLIP$\uparrow$ & Q-Align~\cite{wu2023qalign}$\uparrow$ & PickScore~\cite{Kirstain2023PickaPicAO}$\uparrow$\\
        \midrule
        \rowcolor{gray!10} 
        SDXL~\cite{podell2024sdxl} & 162.18 & 13.23 & \textbf{32.88} & \textbf{4.68} & 0.1468\\
        + FreeU~\cite{si2023freeu} & 167.22 & 12.25 & 31.76 & 4.64 & 0.0967 \\
        + Cosine Sch.~\cite{nichol2021improved} & 182.06 & 11.38 & 30.78 & 2.88 & 0.0376\\
        + Rescaled Sch.~\cite{lin2024common} & 163.29 & 10.88 & 28.80 & 3.25 & 0.0295\\
        \rowcolor{iccvblue!10} 
        + $\varpi(6.0)$ & \textbf{157.47} & \textbf{13.82} & 32.70 & 4.64 & 0.1149 \\
        \rowcolor{iccvblue!10} 
        + $\varpi(-6.0)$ & 170.52 & 13.01 & \underline{32.81} & \underline{4.67} & \textbf{0.1601} \\
        \rowcolor{iccvblue!10} 
        + EXP1 & 173.49 & 12.67 & 32.70 & 4.64 & \underline{0.1578} \\
        \rowcolor{iccvblue!10} 
        + COS1 & \underline{159.87} & \underline{13.25} & 32.64 & 4.60 & 0.0962 \\
     \bottomrule
    \end{tabular}
    \label{tab:quantitative}
    }
    \hfill
    \parbox{.32\linewidth}{
        \centering
        \footnotesize
        \caption{High-Frequency Energy (HFE) of different Omegance settings and their changes w.r.t. SDXL baseline in brackets.}
        \renewcommand\arraystretch{1.13}
        \begin{tabular}{ccc}
        \toprule
         & SSIM$\uparrow$ & HDE (Changes) \\
        \midrule
        \rowcolor{gray!10}
        SDXL~\cite{podell2024sdxl} & 1.0 & 1159.4 (0) \\
        + $\varpi(6.0)$ & 0.8124 & 955.2 (\textcolor{iccvblue}{-204.2}) \\
        + $\varpi(-6.0)$ & 0.7940 & 1680.1 (\textcolor{Red}{+520.7}) \\
        + EXP1 & 0.7087 & 2272.5 (\textcolor{Red}{+1113.1}) \\
        + EXP2 & 0.6926 & 1365.2 (\textcolor{Red}{+205.8}) \\
        + COS1 & 0.8183 & 1004.7 (\textcolor{iccvblue}{-154.7}) \\
        + COS2 & 0.7311 & 612.8 (\textcolor{iccvblue}{-546.6}) \\
     \bottomrule
    \end{tabular}
    
    \label{tab:hfe}
    }
\end{table*}

\begin{table}[h]
\footnotesize
    \centering
    \caption{Win rate of Omegance compared to baselines in granularity control effectiveness and output quality.}
    \begin{tabular}{ccc}
    \toprule
         & Average Rank Accuracy & Output Quality \\
        \midrule
        Omegance & 93.94\% & 81.38\% \\
     \bottomrule
    \end{tabular}
    
    \label{tab:userstudy}
    \vspace{-4mm}
\end{table}

\begin{figure*}[t!]
    \centering
    \vspace{-2mm}
    \includegraphics[width=\linewidth]{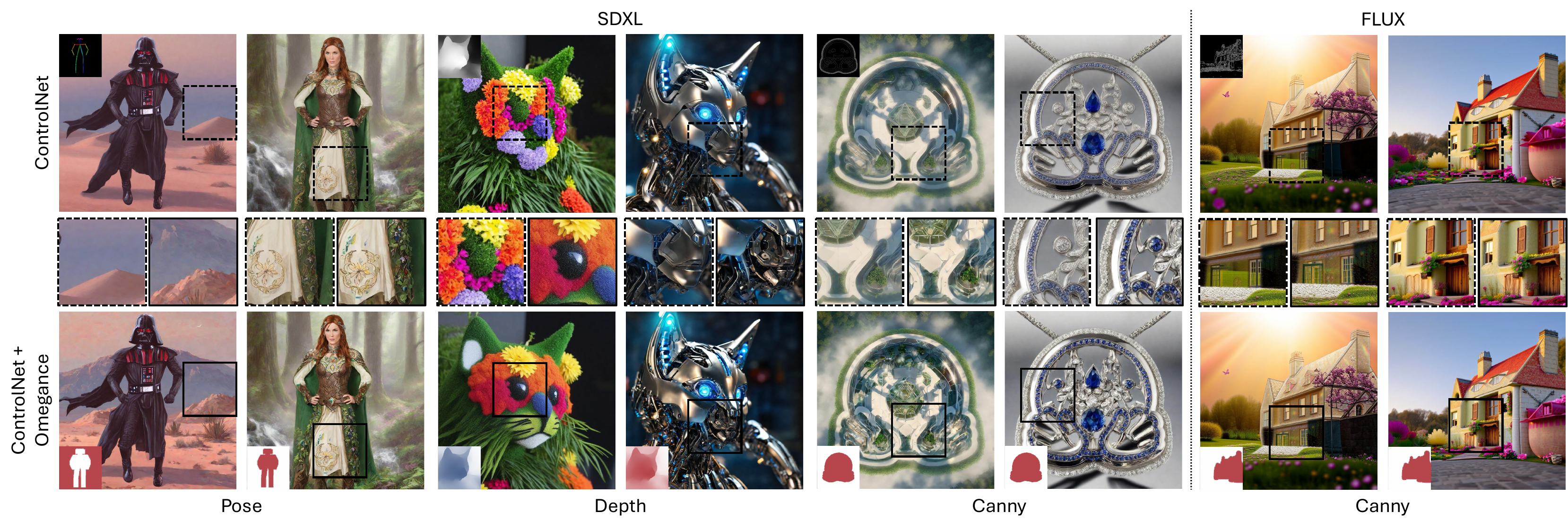}
    \vspace{-6mm}
    \caption{Spatial effects of mask-based Omegance in ControlNet results. \textcolor{black}{Omegance enables spatially controlled granularity adjustments while preserving untouched areas. Mask annotation: red indicates detail enhancement, blue represents detail suppression, and white denotes unchanged regions.}
    }
    \label{fig:controlnet}
    \vspace{-4mm}
\end{figure*}

\noindent \textbf{Global Effect.}
Applying Omegance uniformly across spatial dimensions and consistently over time results in a global granularity change, affecting both the layout and fine details of the output. More qualitative results are shown in Fig.~\ref{fig:global}.

In addition to providing granularity control, Omegance also occasionally enhances the 
generated outputs, as shown in Fig.~\ref{fig:quality}. In lower-quality models, like SDXL~\cite{podell2024sdxl}, Omegance’s detail suppression effectively addresses artifacts in human body parts, particularly in intricate areas like fingers and arms. Meanwhile, for high-quality models like FLUX~\cite{flux_github}, which tend to produce over-smoothed results, Omegance’s detail enhancement improves realism by restoring fine-grained textures and intricate details.


\noindent \textbf{Omega Schedule.} We show two discrete omega schedules in Fig.~\ref{fig:omega-schedule} and their effects in Fig.~\ref{fig:teaser}(c). To further demonstrate the effectiveness of omega schedule in controlling the granularity of the output layout and fine detail simultaneously, we illustrate more continuous schedules in Fig.~\ref{fig:more_schedules}. In the given case, layout complexity is generally reflected by the composition of the chalet and the fine detail richness by the festival decorations and footprints in the snow. \textcolor{black}{Note that these omega schedules are merely illustrative and not exhaustive. One can design own omega schedules by following the effects demonstrated in Fig.~\ref{fig:omega-schedule}.}

\noindent \textbf{Quantitative Results.} 
To assess the effectiveness of Omegance, we conducted experiments using 1,000 randomly sampled prompts from DiffusionDB~\cite{wangDiffusionDBLargescalePrompt2022}, with SDXL as the base model. 
\textcolor{black}{To ensure a comprehensive evaluation, we report Fr\'{e}chet Inception Distance (FID)~\cite{heusel2017fid} and Inception Score (IS)~\cite{salimans2016is} for image quality, CLIP score~\cite{Radford2021clip} for text-image alignment, and Q-Align~\cite{wu2023qalign} and PickScore~\cite{Kirstain2023PickaPicAO} for aesthetics.}
As presented in Tab.~\ref{tab:quantitative}, Omegance (highlighted in blue) outperforms structure-modification methods~\cite{si2023freeu} and scheduler-based approaches~\cite{nichol2021improved, lin2024common} across all key dimensions.
\textcolor{black}{Generally, detail suppression (e.g., $\varpi=6.0$ and COS1) enhances image quality, as indicated by lower FID and higher IS scores compared to the base SDXL model. Conversely, detail enhancement (e.g., $\varpi=-6.0$ and EXP1) maintains CLIP and Q-Align scores on par with SDXL, while significantly improving aesthetic appeal, as reflected in higher PickScore values.}


To validate that Omegance preserves overall image composition, we report structural similarity (SSIM)~\cite{ssim} in Tab.~\ref{tab:hfe}, confirming that it minimally alters the global layout, a finding also supported by qualitative results. 
Analyzing high-frequency components is integral to various image processing applications. Here, we perform frequency-domain analysis to quantify fine-detail changes relative to the base model \textit{(formulation in supp.~Sec.G)}. A higher mean high-frequency energy (HFE) ($+$) indicates enhanced details, while a lower HFE ($-$) reflects detail suppression. As shown in Tab.~\ref{tab:hfe}, Omegance's global effects (2nd and 3rd rows) align with our analysis in Sec.~\ref{sec:omegance}. In addition, the HFE values in rows 4-7 show that the metric conforms closely to the different omega schedules we tested in Fig.~\ref{fig:more_schedules}.
\textcolor{black}{If consistent effect is applied to both early and late inference stages (\eg, EXP1 and COS1), HFE changes align with the effect accordingly. However, when early and late-stage effects differ (\eg, EXP2 and COS2), the resulting HFE changes depend on the severity and duration of each effect across the inference stages. This highlights the flexibility of designing custom omega schedules to selectively control detail granularity in layout structure and texture refinements, based on specific generation requirements.}



\begin{figure}[!t]
    \centering
    \setlength{\abovecaptionskip}{1pt}
    \setlength{\belowcaptionskip}{10pt}
    \renewcommand\arraystretch{1.2}
    \begin{tabular}{@{}c@{\hskip 1pt}c@{\hskip 1pt}c@{}}

        \sffamily \fontsize{6}{6}\selectfont Less Detail & \sffamily \fontsize{6}{6}\selectfont Original & \sffamily \fontsize{6}{6}\selectfont More Detail \\[-2pt]

        \animategraphics[width=0.33\linewidth, autoplay, loop]{6}{figures/gif_imgs/mochi-/}{0}{11} &
        \animategraphics[width=0.33\linewidth, autoplay, loop]{6}{figures/gif_imgs/mochi/}{0}{11} &
        \animategraphics[width=0.33\linewidth, autoplay, loop]{6}{figures/gif_imgs/mochi+/}{0}{11} \\

        \includegraphics[width=0.33\linewidth]{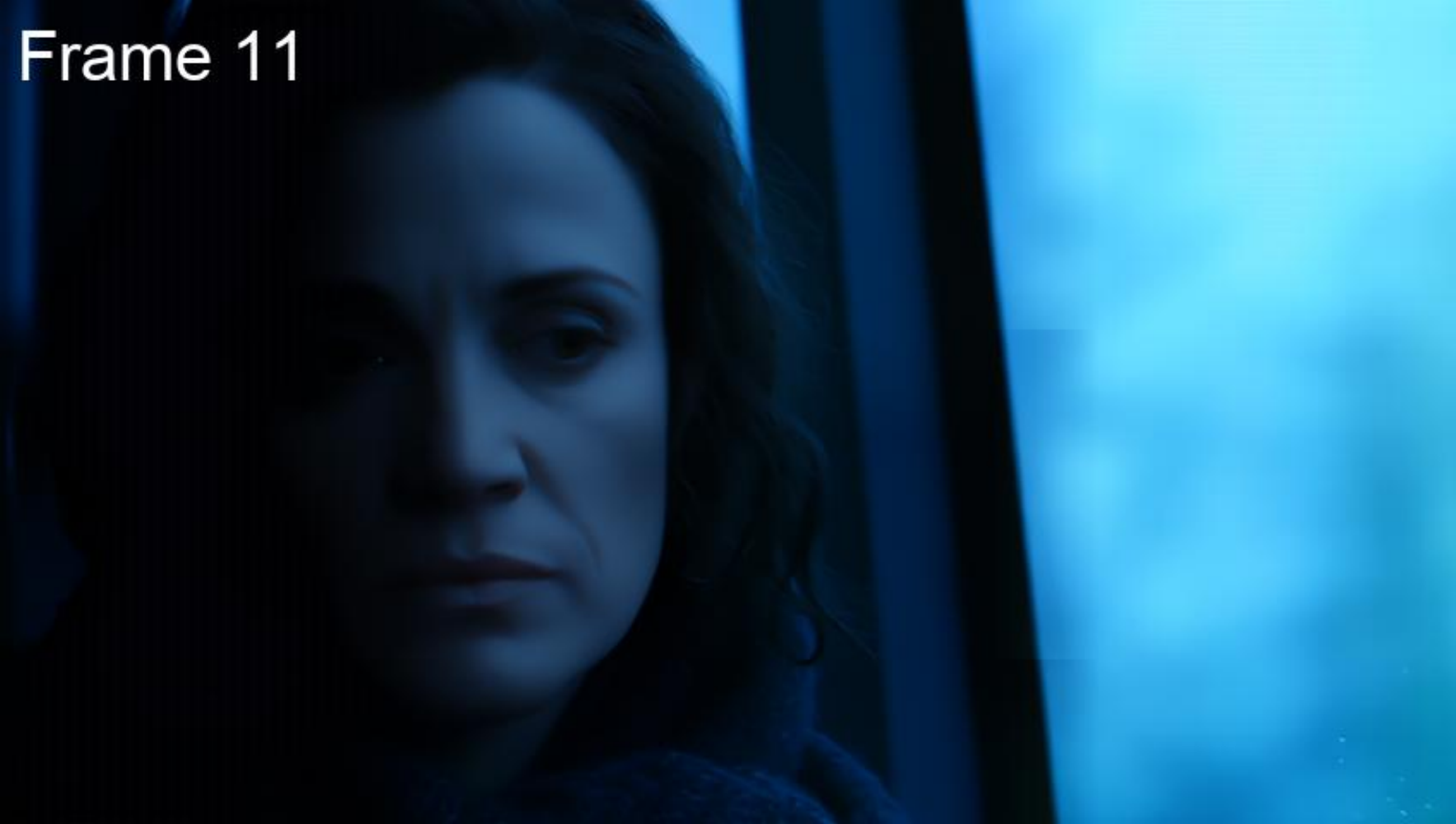} &
        \includegraphics[width=0.33\linewidth]{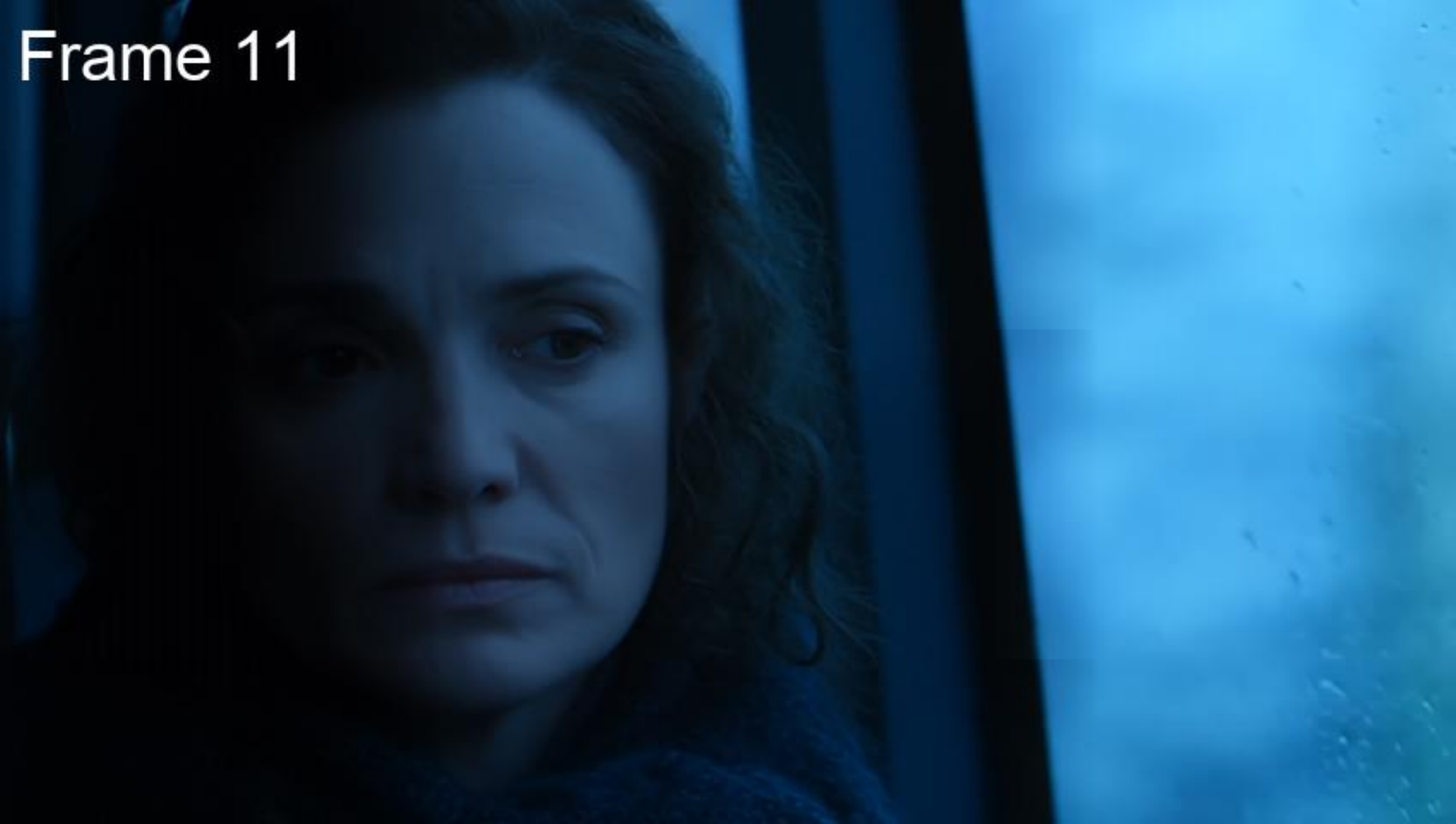} &
        \includegraphics[width=0.33\linewidth]{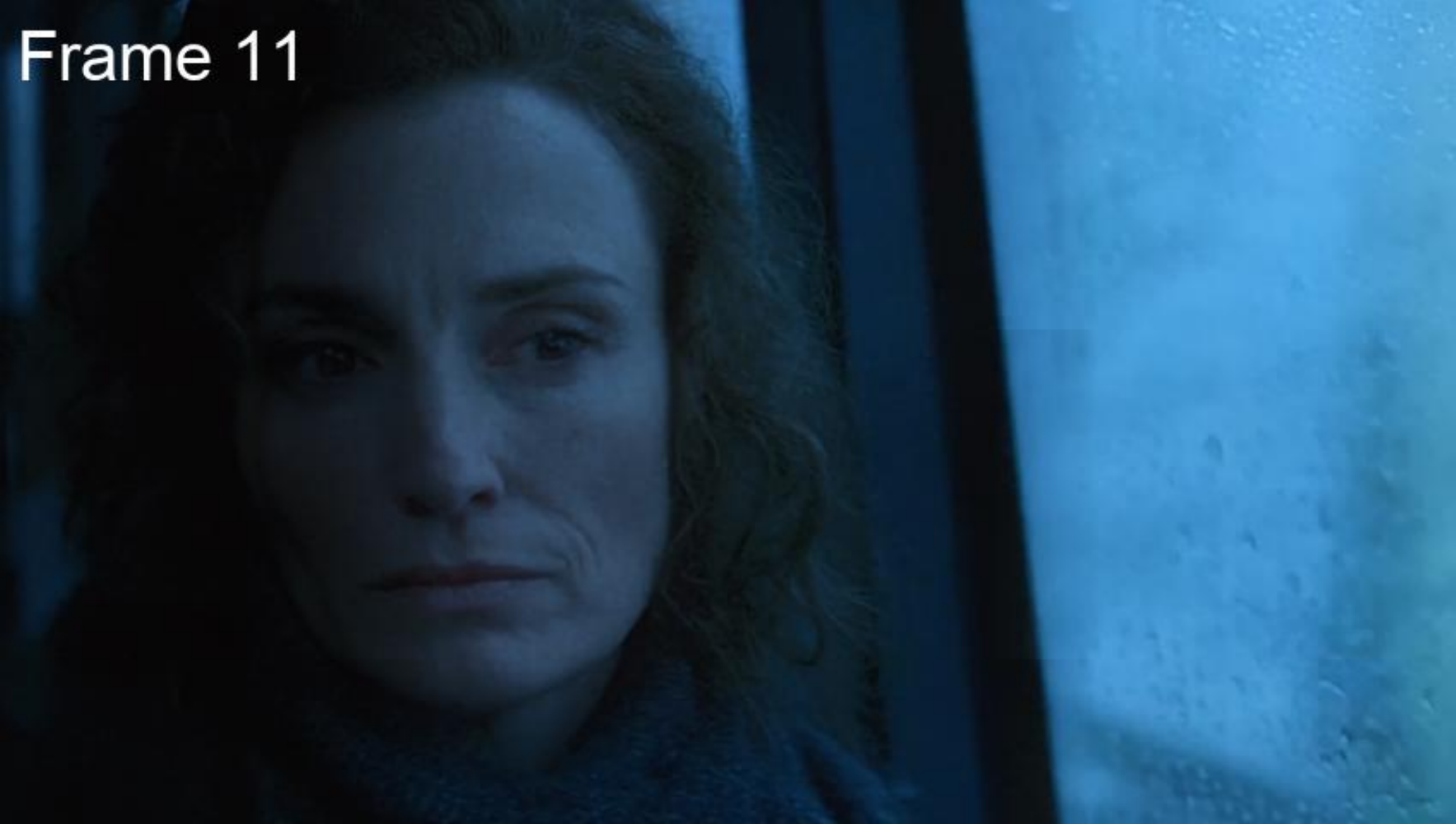} \\[-3pt]

        \multicolumn{3}{c}{\parbox[c]{0.9\linewidth}{
        \centering \sffamily \fontsize{6}{6}\selectfont (a) \textbf{Mochi:} \textit{``Cinematic close-up shot of a sad woman riding a bus in the rain, cool blue tones, sad mood.''}}} \\[6pt]

        \animategraphics[width=0.33\linewidth, autoplay, loop]{6}{figures/gif_imgs/hunyuan-/}{0}{11} &
        \animategraphics[width=0.33\linewidth, autoplay, loop]{6}{figures/gif_imgs/hunyuan/}{0}{11} &
        \animategraphics[width=0.33\linewidth, autoplay, loop]{6}{figures/gif_imgs/hunyuan+/}{0}{11} \\

        \includegraphics[width=0.33\linewidth]{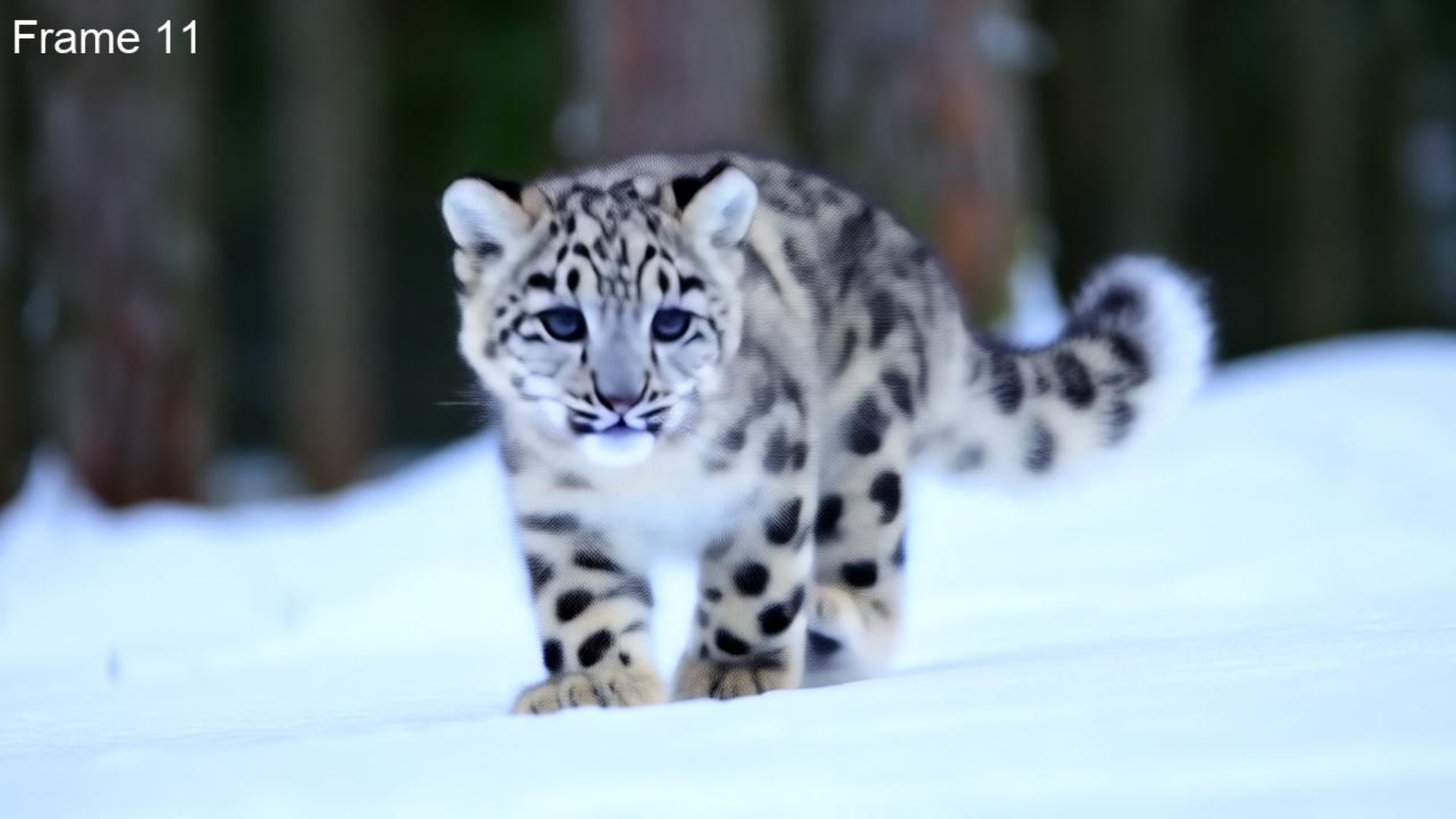} &
        \includegraphics[width=0.33\linewidth]{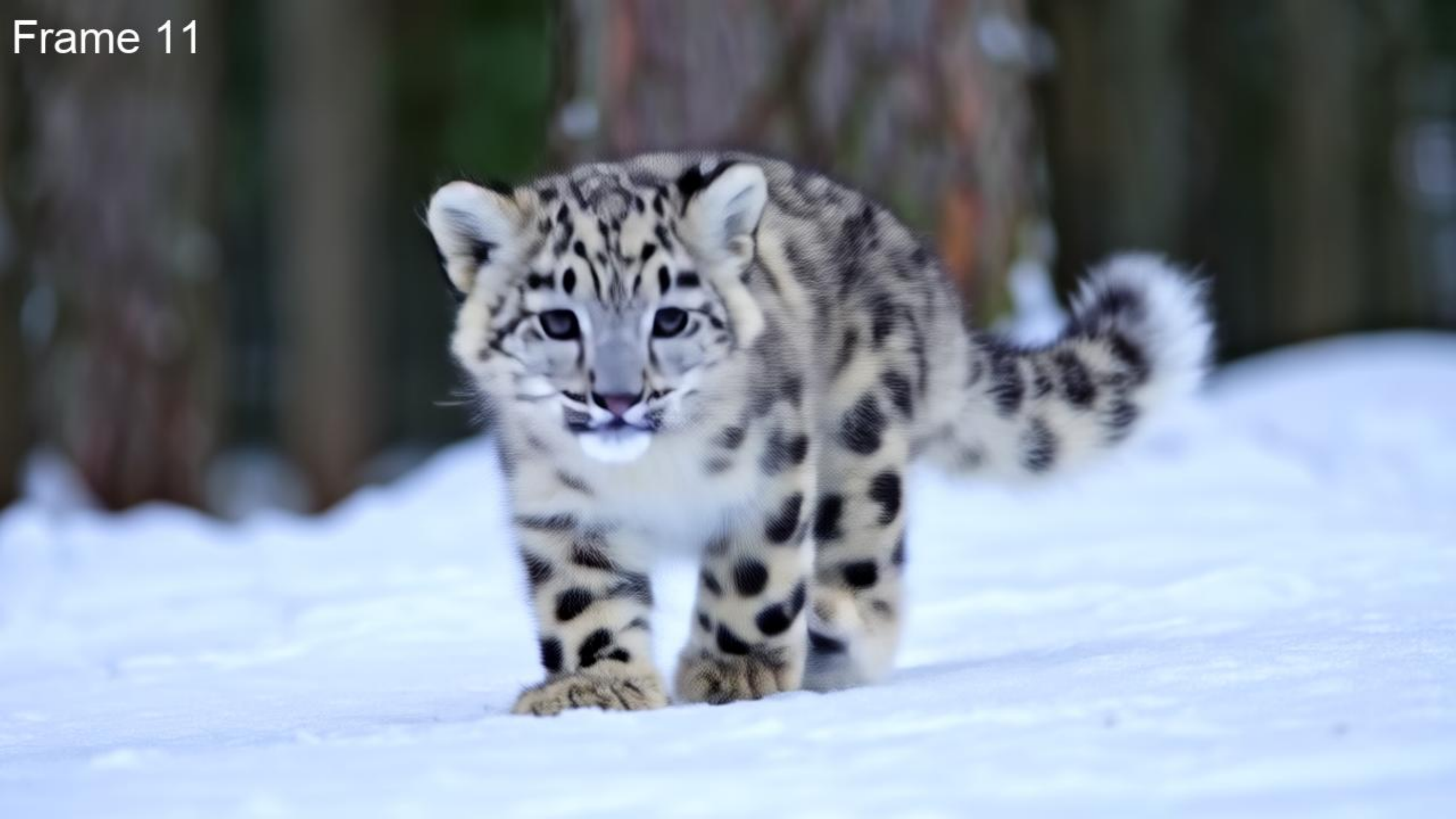} &
        \includegraphics[width=0.33\linewidth]{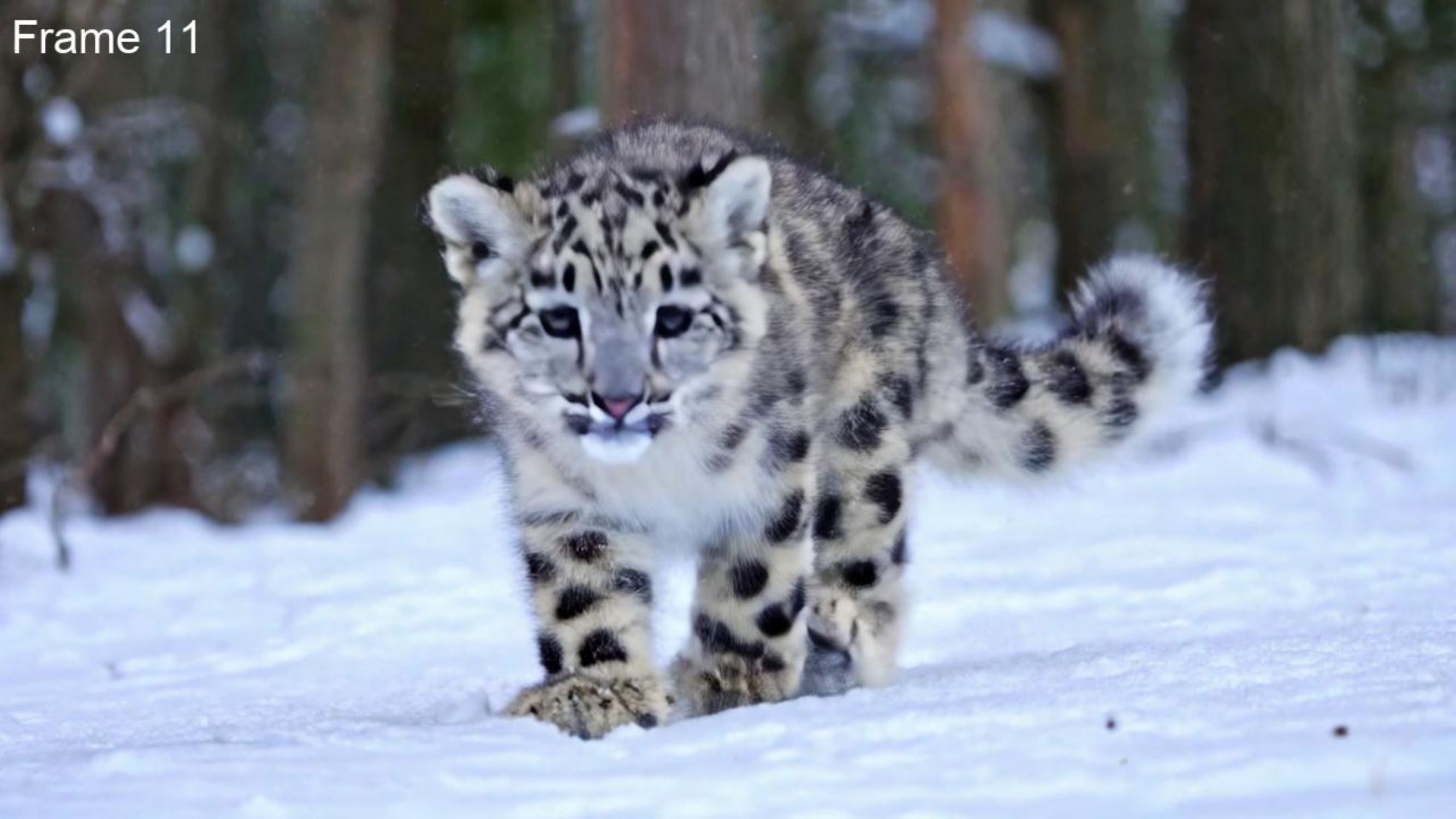} \\[-3pt]

        \multicolumn{3}{c}{\parbox[c]{0.9\linewidth}{
        \centering \sffamily \fontsize{6}{6}\selectfont (b) \textbf{Hunyuan:} \textit{``A cute creature with snow leopard-like fur is walking in winter forest, 3D cartoon style render.''}}} \\[3pt]

    \end{tabular}
    \caption{Effects of Omegance in Text-to-Video results. The employed T2V models and prompts are shown below, with the detail variations on top. Omegance enables granularity control in generated videos while preserving temporal coherence. \textit{(Use Adobe PDF Reader for animated view.)}}
    \vspace{-8mm}
    \label{fig:t2v}
\end{figure}

\noindent{\textbf{User Study.} 
In addition to metric evaluation, we conducted a two-part user study with \textcolor{black}{101} participants to evaluate Omegance's effectiveness in granularity control and impact on output quality \textit{(supp.~Sec.H for details)}. In Part~1, participants are asked to rank three images with/without Omegance based on their granularity. Average rank accuracy reflects the effectiveness of Omegance in granularity control. In Part~2, participants select the higher-quality result from image pairs with/without Omegance, and we report the percentage of votes favoring Omegance or insisting on equal quality. The results in Tab.~\ref{tab:userstudy} demonstrate that Omegance achieves effective granularity control without degrading base model's quality. Notably, 81.38\% of users responded positively to Omegance results, with 67.62\% favoring them and 13.76\% finding them comparable.



\subsection{Image-to-Image Generation}

In image-to-image tasks, prior knowledge of image composition from reference inputs or structural guidance enables the effective use of omega masks to apply Omegance selectively. By assigning specific $\omega$ values to targeted regions, we achieve precise control over texture richness and smoothness, enhancing details in some areas while simplifying others. It is worth noticing that although unselected regions are not explicitly constrained, they experience minimal changes to maintain input consistency, demonstrating the precise region-based control of our omega mask. 

\noindent \textbf{ControlNet.} 
Results of applying omega mask in ControlNet~\cite{zhang2023controlnet} are shown in Fig.~\ref{fig:controlnet}. With ControlNet control signals, we can generate default masks that specify the regions of interest. For the pose signal, which infers the position and pose of the main character, we apply dilated convolution on the skeleton to obtain a default character mask. For a depth signal that conveys foreground and background information, we can use continuous depth values to create depth-aware masks. Alternatively, it is always feasible to generate custom masks from user-provided strokes, allowing for more flexible and intuitive control over detail granularity. We use custom masks for canny signals.


\noindent \textbf{Other tasks.}
More results of SDEdit~\cite{meng2022sdedit}, ReNoise~\cite{garibi2024renoise}, and SDXL-Inpainting~\cite{podell2024sdxl} are shown in supp.~Sec.E.




\subsection{Text-to-Video Generation}

Omegance's granularity control ability also generalizes to text-to-video applications, \eg, Mochi~\cite{genmo2024mochi} and Hunyuan~\cite{kong2024hunyuanvideo}. In Fig.~\ref{fig:t2v}, ``Less Detail'' ($\omega$ increasing) leads to a less complex background and smoother texture, which highlights the main character. On the contrary, ``More Detail'' ($\omega$ decreasing) corresponds to a more complex background and sharper texture, like raindrops on the window in (a) and snow on the ground in (b), leading to more realistic visual results. More text-to-video results of Latte~\cite{ma2024latte} and AnimateDiff~\cite{guo2023animatediff} are shown in supp.~Sec.F.






\section{Conclusion and Limitation}
\label{sec:conclusion}


We introduced Omegance, a simple yet effective single-parameter technique for controlling granularity in diffusion model outputs, enabling fine-grained spatial and temporal adjustments to layout complexity and texture richness. The method is training-free and architecture-agnostic and integrates seamlessly with various diffusion-based tasks.
Extensive experiments demonstrate Omegance's ability to control the level of detail in text-to-image, image-to-image, and text-to-video generation results. 
While Omegance excels at nuanced granularity manipulation and can occasionally correct artifacts or enhance realism, it does not inherently improve the generation quality of the base model, which remains a limitation. 
\textcolor{black}{However, its simplicity is a strength: Unlike existing methods that require model retraining or complex architecture modifications, Omegance leverages a fundamental operation—noise scaling—in a previously unexplored way to enable effective control over generation granularity. By demonstrating its versatility across various tasks, we establish that even a simple modification, when properly applied, can lead to substantial improvements in controllability and user interaction with generative models.}
Nevertheless, we believe this work is valuable for advancing controllable and user-driven content generation, expanding the practical applications of diffusion-based synthesis in real life.


\newpage
\noindent \textbf{Acknowledgement.} This study is supported under the RIE2020 Industry Alignment Fund Industry Collaboration Projects (IAF-ICP) Funding Initiative, as well as cash and in-kind contribution from the industry partner(s).

{
    \small
    \bibliographystyle{ieeenat_fullname}
    \bibliography{main}
}

\end{document}